\documentclass{article}
\usepackage{ijcai17}
\usepackage{times}
\usepackage{graphicx}
\usepackage{amsmath}
\usepackage{amssymb}
\usepackage{url}
\usepackage{caption}
\usepackage{mathrsfs}
\usepackage{algorithm}
\usepackage{dsfont}
\usepackage[pagebackref=false,breaklinks=true,letterpaper=true,colorlinks,urlcolor=blue,citecolor=blue,linkcolor=blue,bookmarks=false]{hyperref}
\usepackage{breakurl}
\usepackage{multirow}
\usepackage{booktabs}

\newcommand{\figdir}{figures}

\renewcommand{\tabcolsep}{.1pt}
\def\avg{\overline}

\def\P{\textbf{P}}
\def\N{\textbf{N}}

\def\A{\textbf{A}}
\def\Y{\textbf{Y}}
\def\y{\textbf{y}}
\def\X{\textbf{X}}
\def\x{\textbf{x}}

\title{Fast Preprocessing for Robust Face Sketch Synthesis}
\author{Yibing Song$^1$, Jiawei Zhang$^1$, Linchao Bao$^2$, and Qingxiong Yang$^3$\\
$^1$City University of Hong Kong\\
$^2$Tencent AI Lab\\
$^3$University of Science and Technology of China\\}

\begin{document}

\maketitle

\begin{abstract}
Exemplar-based face sketch synthesis methods usually meet the challenging problem that input photos are captured in different lighting conditions from training photos. The critical step causing the failure is the search of similar patch candidates for an input photo patch. Conventional illumination invariant patch distances are adopted rather than directly relying on pixel intensity difference, but they will fail when local contrast within a patch changes. In this paper, we propose a fast preprocessing method named Bidirectional Luminance Remapping (BLR), which interactively adjust the lighting of training and input photos. Our method can be directly integrated into state-of-the-art exemplar-based methods to improve their robustness with ignorable computational cost\footnote{Complete experimental results are on the authors' webpage.}.
\end{abstract}

\section{Introduction}

Exemplar-based face sketch synthesis has received much attention in recent years ranging from digital entertainment to law enforcement \cite{liu-ijcai07-bayesian,wang-cvpr12-semi,wang-ijcv14-survey,zhang-tip15-face,Peng-eccv16-lighting,peng-pami16-graphical,wang-tip17-bayesian}.
Typically these methods usually consist of two steps. In the first step, all photos (including a given input photo and all training photos) are divided into local patches, and a K-NN patch search is performed among all training photos for each input photo patch. The second step is to merge the corresponding sketch patches (according to the photo patch search results) into an output sketch image via global optimization \cite{wang-pami2009-face,wei-eccv10-lighting,hao-cvpr12-mwf,wang-nnls13-transductive,zhang-csvt15-face} or local fusion \cite{song-eccv14-sketch}. However, these methods usually fail when input photos are captured differently from training photos which only contain faces in normal lighting. The critical step causing the failure is the search of similar patch candidates for a given input photo patch.

\begin{figure}[!t]
\begin{center}
\begin{tabular}{cccc}
\vspace{-1mm}\includegraphics[width=.245\columnwidth]{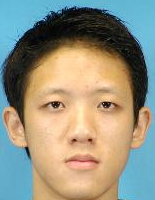}&
\includegraphics[width=.245\columnwidth]{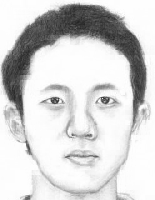}&
\includegraphics[width=.245\columnwidth]{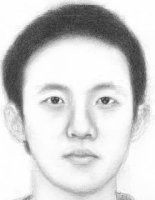}&
\includegraphics[width=.245\columnwidth]{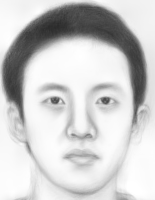}\\
\small{(a) Photo}&\small{(b) MRF}&\small{(c) MWF}&\small{(d) SSD}\\
\vspace{-1mm}\includegraphics[width=.245\columnwidth]{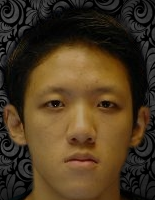}&
\includegraphics[width=.245\columnwidth]{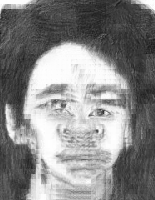}&
\includegraphics[width=.245\columnwidth]{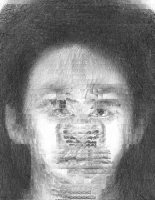}&
\includegraphics[width=.245\columnwidth]{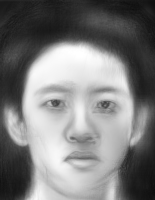}\\
\small{(e) Photo}&\small{(f) MRF}&\small{(g) MWF}&\small{(h) SSD}\\
\vspace{-1mm}\includegraphics[width=.245\columnwidth]{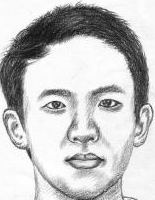}&
\includegraphics[width=.245\columnwidth]{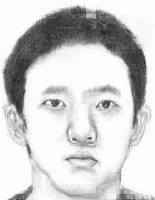}&
\includegraphics[width=.245\columnwidth]{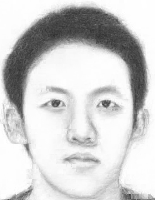}&
\includegraphics[width=.245\columnwidth]{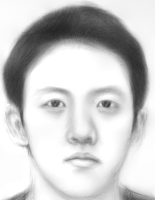}\\
\small{(i) Artist}&\small{(j) Ours+MRF}&\small{(k) Ours+MWF}&\small{(l) Ours+SSD}\\
\end{tabular}
\end{center}
\vspace{-3mm}
\caption{An example of varying lighting conditions. An input photo in (a) is captured in the same condition with training photos. The sketches generated by state-of-the-art MRF, MWF and SSD methods are in (b)-(d). (e) is a synthesized input photo in a different lighting and background condition. (f)-(h) are the results generated by these methods. Our method can be integrated into existing methods to improve the output quality as shown in (j)-(l).}
\label{fig:intro1}
\end{figure}

Most state-of-the-art methods (e.g., MRF \cite{wang-pami2009-face}, MWF \cite{hao-cvpr12-mwf}, and SSD \cite{song-eccv14-sketch}) adopt either $L_1$ or $L_2$ norm based on pixel luminance differences during photo patch search. They perform well on ideal cases where both input and training photos are captured in the same lighting condition. However, for input photos which are captured in different lighting conditions from training photos, these distance metrics often cause incorrect matchings of photo patches and thus lead to erroneous sketch synthesis. Fig. \ref{fig:intro1} shows an example. A direct amending to these methods is to replace the metrics of the pixel luminance difference with illumination invariant ones based on gradient (like DoG \cite{wei-eccv10-lighting}) or correlation (like NCC \cite{szeliski-10-computer}). However, illumination invariant patch distances will fail when local contrast within a patch changes. For example, if the background is brighter than facial skin in input photos while the background is darker than facial skin in training photos, the photo patches near face boundaries are difficult to locate training correspondences (e.g., the left ear region in Fig. \ref{fig:img_global1}(e)). Meanwhile, illumination invariant methods \cite{han-eccv10-lighting,Xie-cvpr08-FaceNormalization} for face recognition are not suitable for face sketch synthesis. They only focus on face region where hair and background are not included.

To enable similar statistics of the face and non-face regions between input and training photos, we propose a novel method, namely \emph{Bidirectional Luminance Remapping} (BLR), to interactively adjust the lighting of both input and training photos. First, the BLR method adapt the lighting of the input photo according to training photos, then it utilizes the offline pre-computed alpha matte information of training photos to recompose them according to the adapted input photo. The advantage of BLR is that it formulate online foreground/background segmentation into offline alpha matting, which enables efficient and accurate patch search. It can be integrated into existing face sketch synthesis methods with ignorable computational cost.

\section{Proposed Algorithm}

In this section, we present the details of how BLR handles lighting variations. Meanwhile, we describe the details for how to integrate BLR into existing methods.

\subsection{Bidirectional Luminance Remapping (BLR)}\label{sec:lr_overall}

\begin{figure}[t]
\begin{center}
\begin{tabular}{cccc}
\includegraphics[width=.245\linewidth]{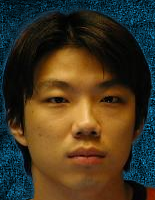}&
\includegraphics[width=.245\linewidth]{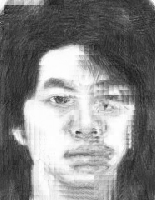}&
\includegraphics[width=.245\linewidth]{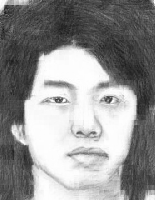}&
\includegraphics[width=.245\linewidth]{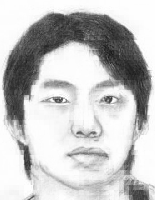}\\
\footnotesize{(a) Photo}&\footnotesize{(b) LR+MRF}
&\footnotesize{(c) ULR+MRF}&\footnotesize{(d) BLR+MRF}\\
\end{tabular}
\end{center}
\vspace{-3mm}
\caption{Improvement with BLR intergration. MRF sketch synthesis method is used in this example. (a) is a challenging input photo captured in dark lighting condition and textured background. (b) is the result with luminance remapping. (c) is the result of the first step of BLR (Sec. \ref{sec:lr_testing}). (d) is the result with BLR.}
\label{fig:algo}
\end{figure}

When an input photo with a different lighting from the training photos is given, a straightforward solution is to perform a global linear luminance remapping (LR) on the input photo to make it contain the same luminance statistics (\emph{e.g.}, mean and variance) with those in training photos \cite{hertzmann-siggraph01-analogy,wei-eccv10-lighting}. However, the global mapping scheme is not applicable for many cases (e.g., when the background has different intensities), and thus, the result is erroneous shown in Fig. \ref{fig:algo}(b).

We now present BLR to make the luminance statistics of the face and non-face regions individually consistent between input and training photos. Each photo consists of the face and non-face regions and the remapping algorithm is performed in two steps. First, we perform a global linear luminance remapping on the input photo according to training photos. Note that this global remapping is only based on the luminance in the face region. It is computed regardless of the non-face region in the training photos, and the non-face region of the input photo can be remapped to arbitrary luminance. In the second step, we adjust the luminance of non-face region in each training photo (using offline pre-computed alpha matte) to make the overall statistics of training photos consistent with those of the input photo obtained in the first step. In this way, the luminance statistics of the face and non-face regions are adjusted similar between input and training photos.

\subsubsection{Luminance Remapping on Input Photo}\label{sec:lr_testing}

We perform luminance remapping on the input photo to enable it contains similar luminance statistics of the face region with those of training photos. The face region is approximately obtained using facial landmarks. We denote $\x$ as the input photo, $\X=a_i\cdot \x+b_i$ as the adapted photo (where $a_i$ and $b_i$ are two scalars), and $\y$ as all the training photos. We denote $\mu_{\x}$, $\mu_{\X}$ and $\mu_{\y}$ as the mean of the face photo(s) $\x$, $\X$ and $\y$, respectively. We denote $\sigma_{\x}$, $\sigma_{\X}$ and $\sigma_{\y}$ as the corresponding standard deviations of $\x$, $\X$ and $\y$, respectively. Our remapping transforms the luminance statistics of the input photo as:
\begin{eqnarray}
    \mu_{\X}&=&a_i\cdot \mu_{\x}+b_i;\\
    \sigma_{\X}^2&=&a_i^2\sigma_{\x}^2;
\end{eqnarray}
The remapping parameters $a_i$ and $b_i$ are computed based on the face region between input and training photos. We denote $\X_{f}$ and $\y_{f}$ as the face region in the adapted photo $\X$ and the training photos $\y$, respectively. We set $\mu_{\X_{f}}=\mu_{\y_{f}}$ and $\sigma_{\X_{f}}=\sigma_{\y_{f}}$ to enable similar luminance statistics on the face region between input and training photos. As a result, parameters $a_i$ and $b_i$ are computed as follows:
\begin{eqnarray}
    a_i&=&\frac{\sigma_{\y_{f}}}{\sigma_{\x_{f}}},\\
    b_i&=&\mu_{\y_{f}}-a_i\cdot\mu_{\x_{f}}.
\end{eqnarray}
We use parameters $a_i$ and $b_i$ to adjust the input photo while not altering training photos at present.

\subsubsection{Luminance Remapping on Training Photos}\label{sec:lr_training}

After conducting luminance remapping in Sec. \ref{sec:lr_testing}, we are confident that the luminance statistics of the face region in adapted input photo $\X$ are similar with those of the training photos. The remaining problem resides in the boundary between the face and non-face regions, which may lead to incorrect patch search and thus the erroneous boundary occurs in the results shown in Fig. \ref{fig:algo}(f). We decompose each training photo into portrait image, non-portrait image and alpha map using matting algorithm \cite{levin-pami08-matting} with manually labeled trimap. The portrait image contains the whole human portrait region while the non-portrait image contains the background region. The non-portrait image is used to approximate the non-face region and the matting operation is done offline. We keep the portrait image fixed and a luminance remapping on the non-portrait image is performed to enable the overall statistics of the training photos similar to those of the adapted input photo obtained in Sec. \ref{sec:lr_testing}.

We denote $\y_p$, $\y_n$ and $\alpha$ as the portrait images, non-portrait images and alpha maps in the training images. So training images $\y$ can be written as
\begin{equation}
    \y=\alpha\cdot \y_p+(1-\alpha)\cdot \y_n.
\end{equation}
We denote $\Y$ as the adapted training images with luminance remapped non-portrait region, then
\begin{eqnarray}\label{eq:Y}
    \Y&=&\alpha\cdot \y_p + (1-\alpha)(a_t\cdot \y_n + b_t)\nonumber\\
    &=&\alpha\cdot \y_p + a_t\cdot (1-\alpha)\cdot \y_n + b_t\cdot (1-\alpha),
\end{eqnarray}
where $a_t$ and $b_t$ are parameters to adjust the non-portrait regions.
We denote $\P = \alpha\cdot \y_p$, $\N=(1-\alpha)\cdot \y_n$, and $\A=(1-\alpha)$. The adapted training images $\Y$ can be written as:
\begin{equation}
    \Y=\P + a\cdot \N + b\cdot \A
\end{equation}
and its mean can be computed as:
\begin{equation}\label{eq:avg_Y}
    \mu_{\Y}=\mu_{\P} + a\cdot \mu_{\N} + b\cdot \mu_{\A}.
\end{equation}
We denote $\avg{\Y}$ as the mean operator on photos $\Y$. So we compute the variance of $\Y$ as:
\begin{eqnarray}\label{eq:sigma_Y}
    \sigma_{\Y}^2&=&\avg{(\Y-\mu_{\Y})^2}\nonumber\\
    &=&\avg{\left((\P - \mu_{\P}) + a_t\cdot (\N-\mu_{\N}) + b_t\cdot (\A-\mu_{\A})\right)^2 }\nonumber\\
    &=&\sigma_{\P}^2 + a_t^2\sigma_{\N}^2 + b_t^2\sigma_{\A}^2 + 2a_t\cdot \sigma_{\P,\N} + 2b_t\cdot \sigma_{\P,\A}\nonumber\\ &&+ 2a_t b_t\cdot \sigma_{\N,\A},
\end{eqnarray}
where $\sigma_{\x,\y}$ corresponds to the covariance between $\x$ and $\y$.

We set $\mu_{\X}=\mu_{\Y}$ and $\sigma_{\X}=\sigma_{\Y}$ to enable the luminance statistics of adapted input photo similar with the adapted training photos. The parameters $a_t$ and $b_t$ can be computed by solving the above two quadratic equations.

In practice, we notice that parameter $b$ is normally small, and thus we can approximate Eq. \eqref{eq:Y} by
\begin{eqnarray}\label{eq:Y2}
    \Y&=&\alpha\cdot \y_p + a_t \cdot \left((1-\alpha)\cdot \y_n\right) + b_t\nonumber\\
    &=&\P + a_t\cdot \N + b_t.
\end{eqnarray}
Then we have
\begin{eqnarray}
    \mu_{\X}=\mu_{\Y}&=&\mu_{\P} + a_t\cdot \mu_{\N} + b_t\label{eq:avg_Y2}\\
    \sigma_{\X}^2=\sigma_{\Y}^2&=&\avg{(\Y-\mu_{\Y})^2}\nonumber\\
    &=&\avg{\left((\P - \mu_{\P}) + a_t\cdot (\N-\mu_{\N})\right)^2 }\nonumber\\
    &=&\sigma_{\N}^2 \cdot a_t^2 + 2\sigma_{\P,\N} \cdot a_t + \sigma_{\P}^2.\label{eq:sigma_Y2}
\end{eqnarray}
Parameter $a_t$ can then be computed by solving the quadratic equation in Eq. \eqref{eq:sigma_Y2} and then used to solve for parameter $b_t$ in Eq. \eqref{eq:avg_Y2}. There will be two possible solutions for the linear transform. To attenuate noise, we choose the positive $a_t$ value that minimizes parameter $b_t$:
\begin{eqnarray}
    a_t&=& \frac{- \sigma_{\P,\N}+\sqrt{ \sigma_{\P,\N}^2- \sigma_{\N}^2 \sigma_{\P}^2+\sigma_{\N}^2\sigma_{\X}^2}}{\sigma_{\N}^2},\\
    b_t&=&\mu_{\X} - \mu_{\P} - a_t\cdot \mu_{\N}.
\end{eqnarray}
After obtaining parameters $a_t$ and $b_t$ we perform remapping on the non-portrait images. Then we recompose training photos using adapted non-portrait image, portrait image, and alpha mat. As a result, we enable similar luminance statistics of face and non-face regions between input and training photos. The photo patch search has been accurate for existing face sketch synthesis methods to synthesize sketches.

\subsection{Practical Issues}\label{sec:practicalconcern}

\subsubsection{Side Lighting}\label{sec:sidelight}
In practice, side lighting may occur in input photos. We use Contrast Limited Adaptive Histogram Equalization (CLAHE) \cite{pizer-1987-clahe} to reduce the effect but find that shadows may still exist around facial components. Then we remap shadow region under the guidance of its symmetric normal lighting region on the face. Specifically, we use landmarks to divide the whole face region into two symmetric parts, i.e, shadow region, and normal lighting region. For each patch in the shadow region, we perform a $1$-NN search in the normal lighting region around the corresponding symmetric position using normalized cross correlation (NCC). Then we remap the luminance of pixels in the shadow region using gamma correction. We denote $T_p$ as a patch centered at a pixel $p$ in the shadow region and $T_{p'}$ as the most similar patch centered at $p'$ in the normal lighting region. The gamma correction can be written as:
\begin{equation}
I_p=I_p^{\mu_{T_p}/\mu_{T_{p'}}}
\end{equation}
where $I_p$ is the luminance of $p$. $\mu_{T_p}$ and $\mu_{T_p'}$ are the mean luminance of patch $T_p$ and $T_p'$, respectively.

\subsubsection{Pose Variance}\label{sec:pose}
In addition to lighting problem, the patch appearance distortion due to pose variations also degrades the quality of selected patch. We precompute the average position of each facial landmark from all training photos to generate a triangulated face template. Given an input photo, we detect its landmarks and compute the local affine transform. Through this transform, input photo is warped to a pose corrected photo, and $K$-NN patch search is then performed between the pose corrected photo and training photos. After sketch synthesis, we warp it back to the original pose.

\subsubsection{Implementation Details}
In our implementation, we precompute the facial landmarks, portrait, and non-portrait images, alpha mattes for all training photos in advance. Given an input photo, we first detect facial landmarks using the algorithm in \cite{Vahid-cvpr14-landmark} and perform local affine transform illustrated in Sec \ref{sec:pose} to warp the input photo into a pose corrected one for further processing. The landmark detection and local affine transform can be conducted in real-time. Second, side lighting is handled as described in Sec \ref{sec:sidelight}. Then  BLR is applied to adapt both input and training photos. After BLR, preprocessed input and training photos can be adopted by existing face sketch synthesis algorithms to synthesize sketch images. Finally, the sketch image is mapped back using local affine transform to yield the final sketch result.

\renewcommand{\tabcolsep}{.1pt}
\begin{figure*}[t]
\begin{center}
\begin{tabular}{cc}
\begin{minipage}[b]{0.01\linewidth}
    \centering
    $\vcenter{\rotatebox{90}{\normalsize MRF}}$
\end{minipage}
\begin{minipage}[t]{0.99\linewidth}
    \centering
    \begin{tabular}{c}
    \includegraphics[width=0.96\linewidth]{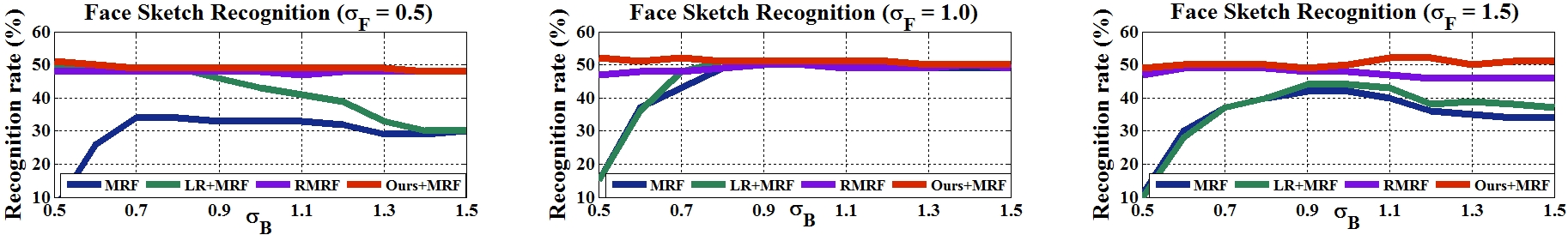}\\
    \end{tabular}
\end{minipage}\\
\begin{minipage}[b]{0.01\linewidth}
    \centering
    $\vcenter{\rotatebox{90}{\normalsize MWF}}$
\end{minipage}
\begin{minipage}[t]{0.99\linewidth}
    \centering
    \begin{tabular}{c}
    \includegraphics[width=0.96\linewidth]{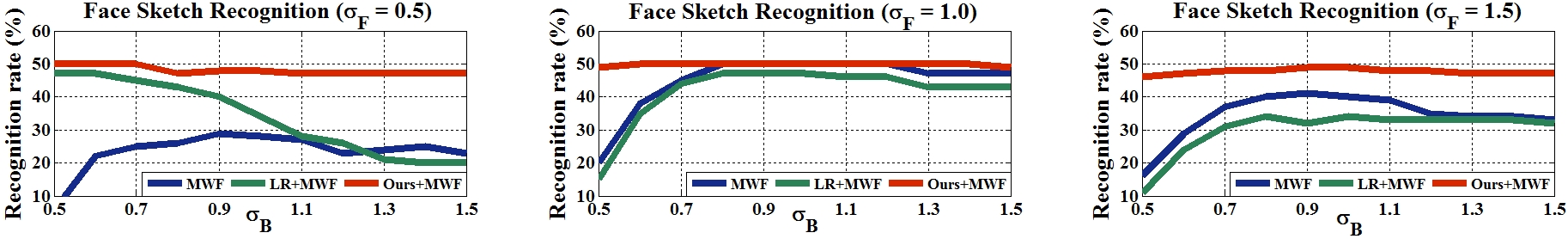}\\
    \end{tabular}
\end{minipage}\\
\begin{minipage}[b]{0.01\linewidth}
    \centering
    $\vcenter{\rotatebox{90}{\normalsize SSD}}$
\end{minipage}
\begin{minipage}[t]{0.99\linewidth}
    \centering
    \begin{tabular}{c}
    \includegraphics[width=0.96\linewidth]{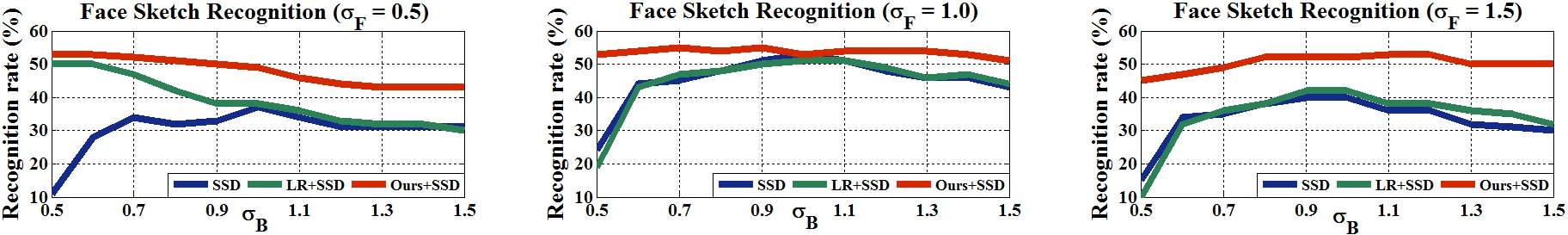}\\
    \end{tabular}
\end{minipage}\\
\begin{minipage}[b]{0.01\linewidth}
    \centering
    $\vcenter{\rotatebox{90}{\normalsize RMRF}}$
\end{minipage}
\begin{minipage}[t]{0.99\linewidth}
    \centering
    \begin{tabular}{c}
    \includegraphics[width=0.96\linewidth]{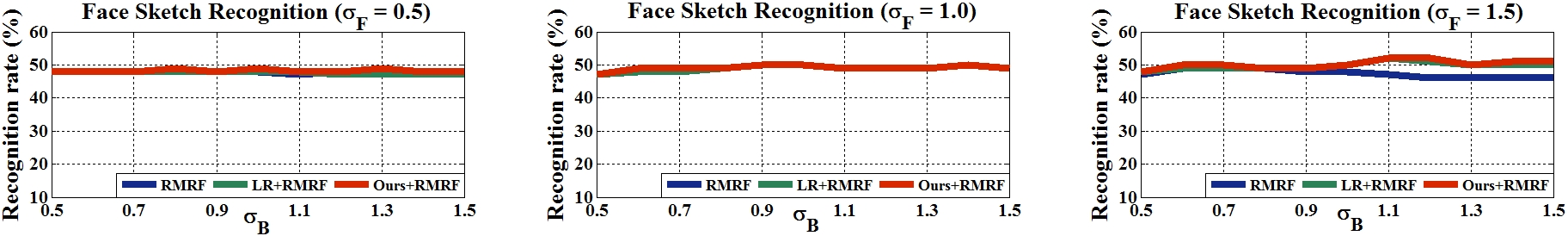}\\
    \end{tabular}
\end{minipage}\\
\end{tabular}
\end{center}
\vspace{-5mm}
\caption{Quantitative evaluation on synthetic CUHK dataset. We use $\sigma_F$ and $\sigma_B$ to adjust the foreground and background lightings of input photos, correspondingly. Our integration improves the robustness of MRF, MWF and SSD regarding to different lightings. It performs favorably against luminance remapping integration and original RMRF method.}
\label{fig:pca_global}
\end{figure*}

\section{Experiments}\label{sec:exp}
We conduct experiments using state-of-the-art face sketch synthesis methods including MRF \cite{wang-pami2009-face}, RMRF \cite{wei-eccv10-lighting}, MWF \cite{hao-cvpr12-mwf} and SSD \cite{song-eccv14-sketch}. The focus is to demonstrate the improvement after integrating BLR into existing methods. The experiments are conducted on the benchmarks including CUHK \cite{wang-pami2009-face}, AR \cite{Mart-TR98-AR}, and FERET datasets \cite{wei-cvpr11-coupled}. The number of photo-sketch pairs for CUHK, AR and FERET are 188, 123 and 1165, respectively. The photos in these three datasets are captured in frontal view and neutral expression. In CUHK dataset the lighting condition is similar for all the photos. In AR dataset the lighting condition is also similar among the photos. However, the lighting condition of CUHK dataset is different from that of AR dataset. For FERET the lighting varies in different photos within this dataset. In addition, we also conduct experiments on the CUHK side lighting and pose variation datasets, which belong to CUHK dataset.

\subsection{Varying Lighting Conditions}\label{sec:explight}

\subsubsection{Synthetic Experiments}\label{sec:syntheticlight}

We first carry out quantitative and qualitative evaluations for synthetic lighting conditions. The synthetic evaluations are conducted on modified CUHK dataset. We split the CUHK dataset into 88 training photo-sketch pairs and 100 input pairs and then generate synthetic input photos in varying lightings as follows. We use matting algorithm \cite{levin-pami08-matting} to divide each input photo into foreground, background and alpha matte images. Then we separately adjust the luminance of foreground and background using two scalars (i.e., $\sigma_F$ and $\sigma_B$). The luminance values of all foreground pixels are multiplied by $\sigma_F$ and those of background are multiplied by $\sigma_B$. Then we combine adjusted foreground and background images with alpha matte to generate synthetic input photos.

We compare our method with baseline luminance remapping (LR) \cite{hertzmann-siggraph01-analogy} preprocessing. Note that RMRF is specially designed for improving the robustness of MRF, we compare with RMRF when evaluating the improvement on MRF. In addition, RMRF can be treated as an independent method which can be integrated with our algorithm. So we first evaluate the original performance of MRF, MWF, SSD and RMRF. Then we compare the improvements of the LR and our integration of these methods.

Quantitative evaluation of face sketch synthesis methods can be conducted through face sketch recognition as suggested in \cite{wang-pami2009-face}. For each input photo, the synthesized sketch should be matched to the corresponding sketch drawn by the artist. If an algorithm achieves higher sketch recognition rates, it suggests that this method is more robust to synthesize sketches. Fig. \ref{fig:pca_global} shows the performance of quantitative evaluation. The foreground of input photos is adjusted by three values of $\sigma_F$, i.e., 0.5, 1.0, and 1.5. These values simulate the dark, normal, and bright foreground of input photos, respectively. For each $\sigma_F$ we adjust $\sigma_B$ from 0.5 to 1.5 incremented by 0.1, which simulates the varying background lightings. The results show that MRF, MWF, and SSD are not robust to synthesize sketches from photos captured in different lightings. Due to its global normalization scheme, LR preprocessing cannot robustly handle all lighting conditions. Our algorithm can consistently improve the performance of existing methods. Compared with RMRF, our algorithm is more robust against extreme cases (the first row of Fig. \ref{fig:pca_global}). Moreover, our algorithm can be integrated with RMRF to improve its robustness (the last row of Fig. \ref{fig:pca_global}).

\begin{figure}[t]
\begin{center}
\begin{tabular}{cccc}
\vspace{-1mm}\includegraphics[width=.245\columnwidth]{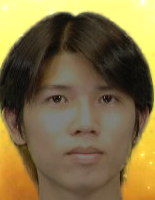}&
\includegraphics[width=.245\columnwidth]{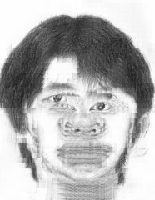}&
\includegraphics[width=.245\columnwidth]{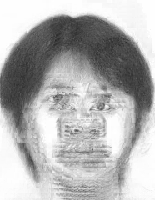}&
\includegraphics[width=.245\columnwidth]{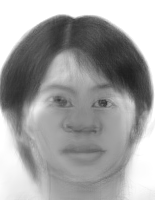}\\
\small{(a) Photo}&\small{(b) MRF}&\small{(c) MWF}&\small{(d) SSD}\\
\vspace{-1mm}\includegraphics[width=.24\columnwidth]{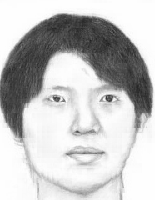}&
\includegraphics[width=.245\columnwidth]{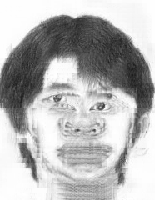}&
\includegraphics[width=.245\columnwidth]{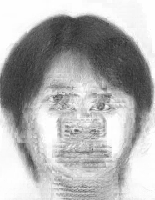}&
\includegraphics[width=.245\columnwidth]{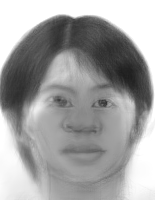}\\
\small{(e) RMRF}&\small{(f) LR+MRF}&\small{(g) LR+MWF}&\small{(h) LR+SSD}\\
\vspace{-1mm}\includegraphics[width=.24\columnwidth]{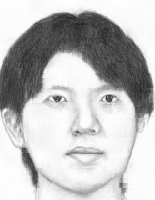}&
\includegraphics[width=.245\columnwidth]{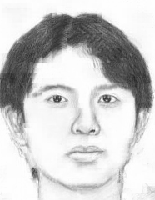}&
\includegraphics[width=.245\columnwidth]{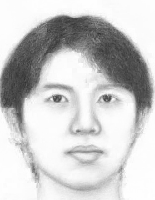}&
\includegraphics[width=.245\columnwidth]{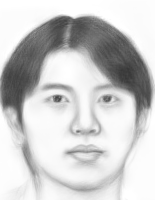}\\
\small{(i)Ours+RMRF}&\small{(j)Ours+MRF}&\small{(k)Ours+MWF}&\small{(l)Ours+SSD}\\
\end{tabular}
\end{center}
\vspace{-4mm}
\caption{An example of synthetic lighting experiments. (a) is the input photo consists of dark foreground and bright background. (b)-(e) are the results of existing methods. (f)-(h) are the results of improved existing methods with luminance remapping integration. (i)-(l) are the results of improved existing methods with our integration.}
\label{fig:img_global1}
\end{figure}

Fig. \ref{fig:img_global1} shows one example of the visual comparison for the synthetic evaluation. The input photo consists of dark foreground and bright background. As the foreground differs from training photos patch candidates can not be correctly matched, which results in blurry and artifacts as shown in (b)-(d). LR based on global luminance statistics fails to correct the lightings and thus produces erroneous results as shown in (f)-(h). In comparison, BLR adapts both input and training photos to enable more accurate patch search in the face and non-face regions. As a result, the accuracy of $K$-NN patch searching is improved and the obtained sketch results achieve ideal performance as shown in (i)-(l). Meanwhile, the local contrast within photo patch is reduced through our integration and thus the result in (i) is improved around face boundary.

\subsubsection{Cross-Dataset Experiments}
\begin{figure}
\begin{center}
\begin{tabular}{cccc}
\vspace{-1mm}\includegraphics[width=.245\columnwidth]{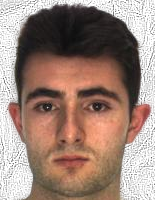}&
\includegraphics[width=.245\columnwidth]{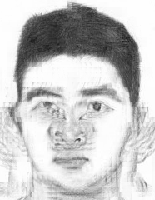}&
\includegraphics[width=.245\columnwidth]{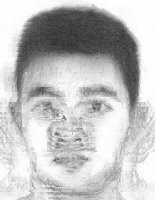}&
\includegraphics[width=.245\columnwidth]{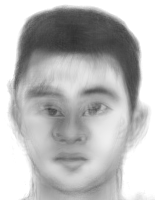}\\
\small{(a) Photo}&\small{(b) MRF}&\small{(c) MWF}&\small{(d) SSD}\\
\vspace{-1mm}\includegraphics[width=.245\columnwidth]{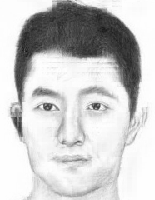}&
\includegraphics[width=.245\columnwidth]{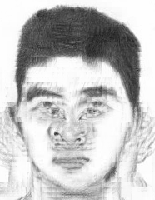}&
\includegraphics[width=.245\columnwidth]{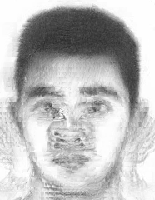}&
\includegraphics[width=.245\columnwidth]{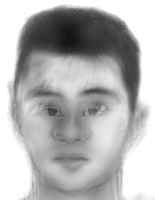}\\
\small{(e) RMRF}&\small{(f) LR+MRF}&\small{(g) LR+MWF}&\small{(h) LR+SSD}\\
\vspace{-1mm}\includegraphics[width=.245\columnwidth]{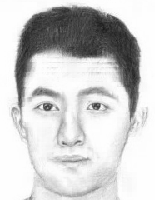}&
\includegraphics[width=.245\columnwidth]{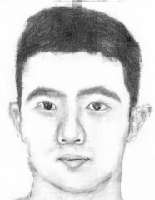}&
\includegraphics[width=.245\columnwidth]{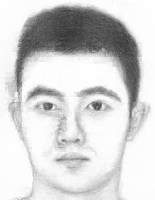}&
\includegraphics[width=.245\columnwidth]{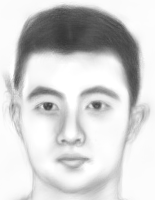}\\
\small{(i)Ours+RMRF}&\small{(j)Ours+MRF}&\small{(k)Ours+MWF}&\small{(l)Ours+SSD}\\
\end{tabular}
\end{center}
\vspace{-4mm}
\caption{An example of cross-dataset experiments (CUHK as training while AR as input). (a) is an input photo. (b)-(l) are with the same meaning as Fig. \ref{fig:img_global1}.}
\label{fig:img_arcuhk}
\end{figure}

We notice that CUHK and AR datasets are captured in different lightings. Thus we evaluate the robustness of BLR using CUHK as training and AR as input and vice versa. Fig. \ref{fig:img_arcuhk} shows the visual comparison where BLR can consistently improve existing methods. Although ethnic facial difference exists between two datasets, BLR can still robustify sketch synthesis of existing methods.

\subsubsection{Real Lighting Experiments}
\begin{figure}
\begin{center}
\begin{tabular}{cccc}
\vspace{-1mm}\includegraphics[width=.245\columnwidth]{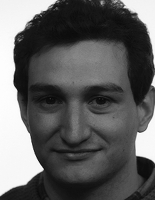}&
\includegraphics[width=.245\columnwidth]{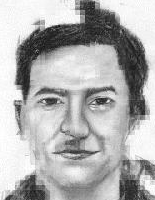}&
\includegraphics[width=.245\columnwidth]{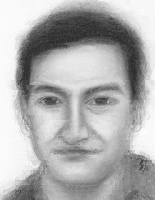}&
\includegraphics[width=.245\columnwidth]{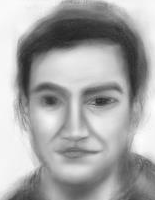}\\
\small{(a) Photo}&\small{(b) MRF}&\small{(c) MWF}&\small{(d) SSD}\\
\vspace{-1mm}\includegraphics[width=.245\columnwidth]{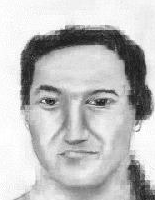}&
\includegraphics[width=.245\columnwidth]{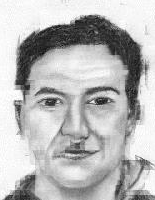}&
\includegraphics[width=.245\columnwidth]{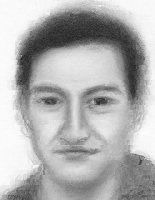}&
\includegraphics[width=.245\columnwidth]{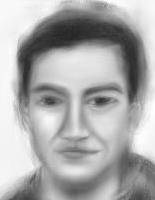}\\
\small{(e) RMRF}&\small{(f) LR+MRF}&\small{(g) LR+MWF}&\small{(h) LR+SSD}\\
\vspace{-1mm}\includegraphics[width=.245\columnwidth]{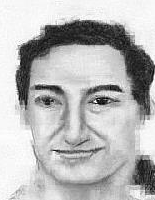}&
\includegraphics[width=.245\columnwidth]{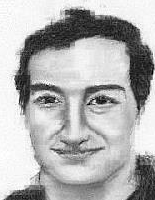}&
\includegraphics[width=.245\columnwidth]{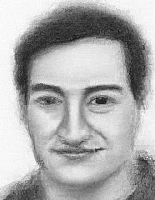}&
\includegraphics[width=.245\columnwidth]{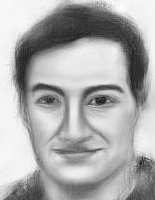}\\
\small{(i)Ours+RMRF}&\small{(j)Ours+MRF}&\small{(k)Ours+MWF}&\small{(l)Ours+SSD}\\
\end{tabular}
\end{center}
\vspace{-4.5mm}
\caption{An example of experiments on FERET dataset. (a) is an input photo. (b)-(l) are with the same meaning as Fig. \ref{fig:img_global1}.}
\label{fig:img_feret2}
\end{figure}

We conduct an evaluation of BLR on FERET dataset. Different from the previous two datasets FERET contains photos captured in real world varying lighting conditions. We randomly select 100 photo-sketch pairs as training and use the remaining 1065 pairs as input. Fig. \ref{fig:img_feret2} shows one example of the visual evaluation. The lighting is different in both foreground and background regions, which leads to artifacts on the synthesized sketches of existing methods. Through our integration, the statistics of the face and non face regions are adjusted similarly among input and training photos. It enables existing methods to robustify sketch synthesis.

\subsubsection{Side Lighting Experiments}

\begin{figure}[!ht]
\begin{center}
\begin{tabular}{cccc}
\vspace{-1mm}\includegraphics[width=.245\columnwidth]{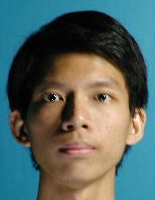}&
\includegraphics[width=.245\columnwidth]{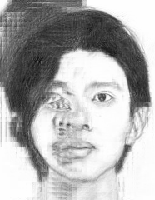}&
\includegraphics[width=.245\columnwidth]{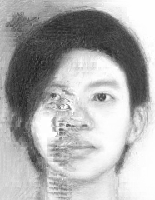}&
\includegraphics[width=.245\columnwidth]{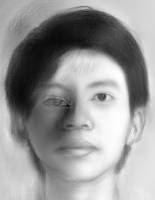}\\
\small{(a) Photo}&\small{(b) MRF}&\small{(c) MWF}&\small{(d) SSD}\\
\vspace{-1mm}\includegraphics[width=.245\columnwidth]{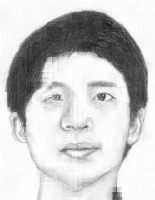}&
\includegraphics[width=.245\columnwidth]{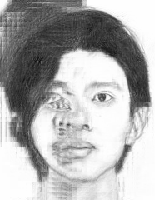}&
\includegraphics[width=.245\columnwidth]{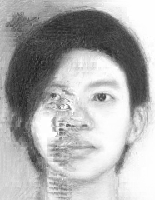}&
\includegraphics[width=.245\columnwidth]{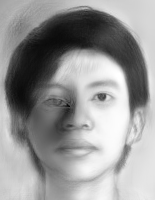}\\
\small{(e) RMRF}&\small{(f) LR+MRF}&\small{(g) LR+MWF}&\small{(h) LR+SSD}\\
\vspace{-1mm}\includegraphics[width=.245\columnwidth]{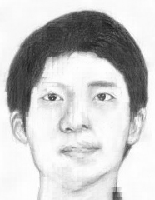}&
\includegraphics[width=.245\columnwidth]{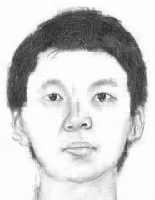}&
\includegraphics[width=.245\columnwidth]{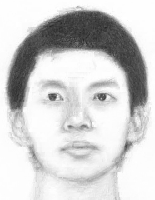}&
\includegraphics[width=.245\columnwidth]{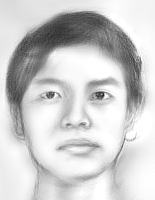}\\
\small{(i)Ours+RMRF}&\small{(j)Ours+MRF}&\small{(k)Ours+MWF}&\small{(l)Ours+SSD}\\
\end{tabular}
\end{center}
\vspace{-4.5mm}
\caption{An example of side lighting experiments. (a) is an input photo. (b)-(l) are with the same meaning as Fig. \ref{fig:img_global1}.}
\label{fig:img_local}
\end{figure}

We conduct experiments on CUHK side lighting dataset \cite{wei-eccv10-lighting} which contains two different types of side lighting (dark left / dark right) photos for each subject. As the input photo contains shadows in the facial region shown in Fig. \ref{fig:img_local}, existing methods cannot find correctly matched photo patches around these shadow regions. It leads to blur and artifacts shown in (b)-(d). In comparison, Our method can locally adjust input photo to make an improvement.

\subsection{Varying Poses}\label{sec:varyposeexp}

\begin{figure}[!h]
\begin{center}
\begin{tabular}{cccc}
\vspace{-1mm}\includegraphics[width=.245\columnwidth]{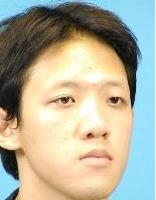}&
\includegraphics[width=.245\columnwidth]{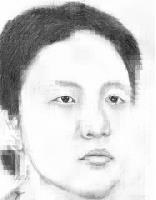}&
\includegraphics[width=.245\columnwidth]{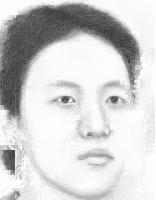}&
\includegraphics[width=.245\columnwidth]{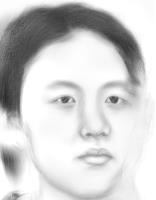}\\
\footnotesize{(a) Photo}&\footnotesize{(b) MRF}&\footnotesize{(c) MWF}&\footnotesize{(d) SSD}\\
\vspace{-1mm}\includegraphics[width=.245\columnwidth]{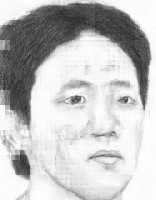}&
\includegraphics[width=.245\columnwidth]{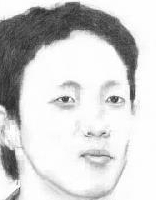}&
\includegraphics[width=.245\columnwidth]{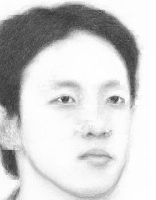}&
\includegraphics[width=.245\columnwidth]{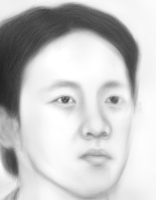}\\
\footnotesize{(e) RMRF-ext}&\footnotesize{(f) MRF-ext}&\footnotesize{(g) MWF-ext}&\footnotesize{(h) SSD-ext}\\
\vspace{-1mm}\includegraphics[width=.245\columnwidth]{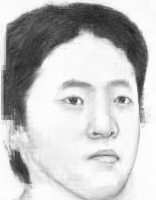}&
\includegraphics[width=.245\columnwidth]{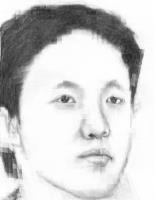}&
\includegraphics[width=.245\columnwidth]{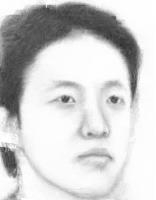}&
\includegraphics[width=.245\columnwidth]{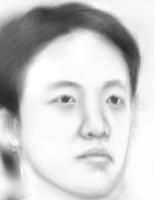}\\
\footnotesize{(i)Ours+RMRF}&\footnotesize{(j)Ours+MRF}&\footnotesize{(k)Ours+MWF}&\footnotesize{(l)Ours+SSD}\\
\end{tabular}
\end{center}
\vspace{-4.5mm}
\caption{An example of varying pose experiments. (a) is an input photo. (b)-(d) are the synthesized sketches. (e)-(h) are the results synthesized with extended search range. (i)-(l) are the sketches synthesized with our integration.}
\label{fig:img_pose}
\end{figure}

We perform experiments on CUHK pose variation dataset \cite{wei-eccv10-lighting} where subjects are in varying poses. Note that some methods \cite{song-eccv14-sketch,hao-cvpr12-mwf} tend to increase search range for handling varying poses. Thus we also compare BLR with existing methods using extended search range. Fig. \ref{fig:img_pose} shows an example of the visual evaluation result. Our algorithm favorably improves the robustness of existing methods.

\def\pp{\hspace{2mm}}
\renewcommand{\tabcolsep}{7pt}
\begin{table}
\vspace{-3mm}\caption{Runtime (seconds) for a CUHK input image.}
\centering
\vspace{-3mm}
       \begin{tabular}{ccccc}
        \toprule
        \scriptsize{}&\scriptsize{MRF}\pp&\pp\scriptsize{MWF}\pp&\pp\scriptsize{RMRF}&\pp\scriptsize{SSD}\\
        \midrule
        \scriptsize{Original}&\pp\scriptsize{38.4}&\pp\scriptsize{35.6}\pp&\pp\scriptsize{88.2}&\pp\scriptsize{4.5}\\
        \scriptsize{Original (ext.)$^\dag$}&\pp\scriptsize{94.5}&\pp\scriptsize{93.8}\pp&\pp\scriptsize{252.3}&\pp\scriptsize{13.4}\\
        \scriptsize{Original + Ours}&\pp\scriptsize{38.6}&\pp\scriptsize{35.9}\pp&\pp\scriptsize{93.5}&\pp\scriptsize{4.7}\\
        \bottomrule
        \multicolumn{5}{l}{\scriptsize{$^\dag$With extended search range (see Sec. \ref{sec:varyposeexp}).}}
       \end{tabular}
\label{tab:time}
\end{table}

\subsection{Computational Cost}
Table \ref{tab:time} shows the runtime of existing methods to process a CUHK input image (obtained from a 3.4GHz Intel i7 CPU). It shows that the additional computation cost brought by BLR is ignorable compared with the original time cost of existing methods. Note that the reason why RMRF needs more additional computational cost is that we need to extract features online of the recomposed training photos.

\section{Concluding Remarks}
We propose BLR, which interactively adjusts the lighting of training and input photos. It moves online face image segmentation to offline using human supervised alpha matting. The experiments demonstrate that BLR improves the robustness of existing methods with ignorable computational cost.

\clearpage
\bibliographystyle{named}
\bibliography{ref}

\end{document}